\documentclass[letterpaper]{article} 
\usepackage{aaai2026}  
\usepackage{times}  
\usepackage{helvet}  
\usepackage{courier}  
\usepackage[hyphens]{url}  
\usepackage{graphicx} 
\usepackage{natbib}  
\usepackage{caption} 
\usepackage{algorithm}
\usepackage{algorithmic}
\usepackage{newtxtext,newtxmath}
\usepackage{makecell}

\usepackage{newfloat}

\usepackage{amsthm}
\newtheorem{theorem}{Theorem}
\usepackage{amsmath}
\usepackage{booktabs}
\usepackage{listings}
\DeclareCaptionStyle{ruled}{labelfont=normalfont,labelsep=colon,strut=off} 
\floatstyle{ruled}
\newfloat{listing}{tb}{lst}{}
\floatname{listing}{Listing}
\title{Ambiguity-aware Truncated Flow Matching for Ambiguous Medical Image Segmentation}
\author{
Fanding Li\textsuperscript{\rm 1}, Xiangyu Li\textsuperscript{\rm 1}\thanks{Corresponding Author}, Xianghe Su\textsuperscript{\rm 1}, Xingyu Qiu\textsuperscript{\rm 1}, Suyu Dong\textsuperscript{\rm 2}, Wei Wang\textsuperscript{\rm 3}, Kuanquan Wang\textsuperscript{\rm 1}, Gongning Luo\textsuperscript{\rm 1}, Shuo Li\textsuperscript{\rm 4,5}
}
\affiliations{
\textsuperscript{\rm 1}Faculty of Computing, Harbin Institute of Technology, Harbin, China\\
\textsuperscript{\rm 2}College of Computer and Control Engineering, Northeast Forestry University, Harbin, China\\
\textsuperscript{\rm 3}Faculty of Computing, Harbin Institute of Technology, Shenzhen, China\\
\textsuperscript{\rm 4}Department of Computer and Data Science, Case Western Reserve University, Cleveland, Ohio 44106, United States\\
\textsuperscript{\rm 5}Department of Biomedical Engineering, Case Western Reserve University, Cleveland, Ohio 44106, United States\\

lixiangyu@hit.edu.cn\\
}

\usepackage{bibentry}

\begin{document}

\maketitle

\begin{abstract}
	A simultaneous enhancement of accuracy and diversity of predictions remains a challenge in ambiguous medical image segmentation (AMIS) due to the inherent trade-offs. While truncated diffusion probabilistic models (TDPMs) hold strong potential with a paradigm optimization, existing TDPMs suffer from entangled accuracy and diversity of predictions with insufficient fidelity and plausibility. To address the aforementioned challenges, we propose Ambiguity-aware Truncated Flow Matching (ATFM), which introduces a novel inference paradigm and dedicated model components. Firstly, we propose Data-Hierarchical Inference, a redefinition of AMIS-specific inference paradigm, which enhances accuracy and diversity at data-distribution and data-sample level, respectively, for an effective disentanglement. Secondly, Gaussian Truncation Representation (GTR) is introduced to enhance both fidelity of predictions and reliability of truncation distribution, by explicitly modeling it as a Gaussian distribution at $T_{\text{trunc}}$ instead of using sampling-based approximations. Thirdly, Segmentation Flow Matching (SFM) is proposed to enhance the plausibility of diverse predictions by extending semantic-aware flow transformation in Flow Matching (FM). Comprehensive evaluations on LIDC and ISIC3 datasets demonstrate that ATFM outperforms SOTA methods and simultaneously achieves a more efficient inference. ATFM improves GED and HM-IoU by up to $12\%$ and $7.3\%$ compared to advanced methods.
\end{abstract}

\begin{links}
	\link{Code}{https://github.com/PerceptionComputingLab/ATFM}
	\link{Extended version}{}This  is  the  extended  version  of  AAAI  paper with full appendixes
\end{links}

\section{Introduction}

Generating a series of predictions with high accuracy and diversity to estimate the distribution of annotation space is of significant importance in ambiguous medical image segmentation (AMIS) \cite{5,4}. High diversity reflects the inherent ambiguity present in medical images and high accuracy is critical for supporting dependable clinical decision-making, making both essential components of a reliable AMIS framework \cite{7,6}.

However, simultaneously improving prediction accuracy and diversity remains challenging in AMIS due to the inherent trade-off between these objectives in existing methods. Stochastic approaches \cite{7,1} enhance diversity at the expense of the more important accuracy, yielding low-confidence diagnoses. Zhang et al. \cite{6} are able to regulate this trade-off but cannot enhance both properties simultaneously. Multi-rater-aware techniques \cite{17,34} improve accuracy and diversity by modeling annotators' labeling styles, yet their annotator-centric design inherently suppresses low-frequency modes, degrading segmentation quality. Broader application of these methods is still constrained by the inherent trade-off between entangled accuracy and diversity among predictions.

\begin{figure}
	\centering 	 	
	\includegraphics[width = 1\linewidth]{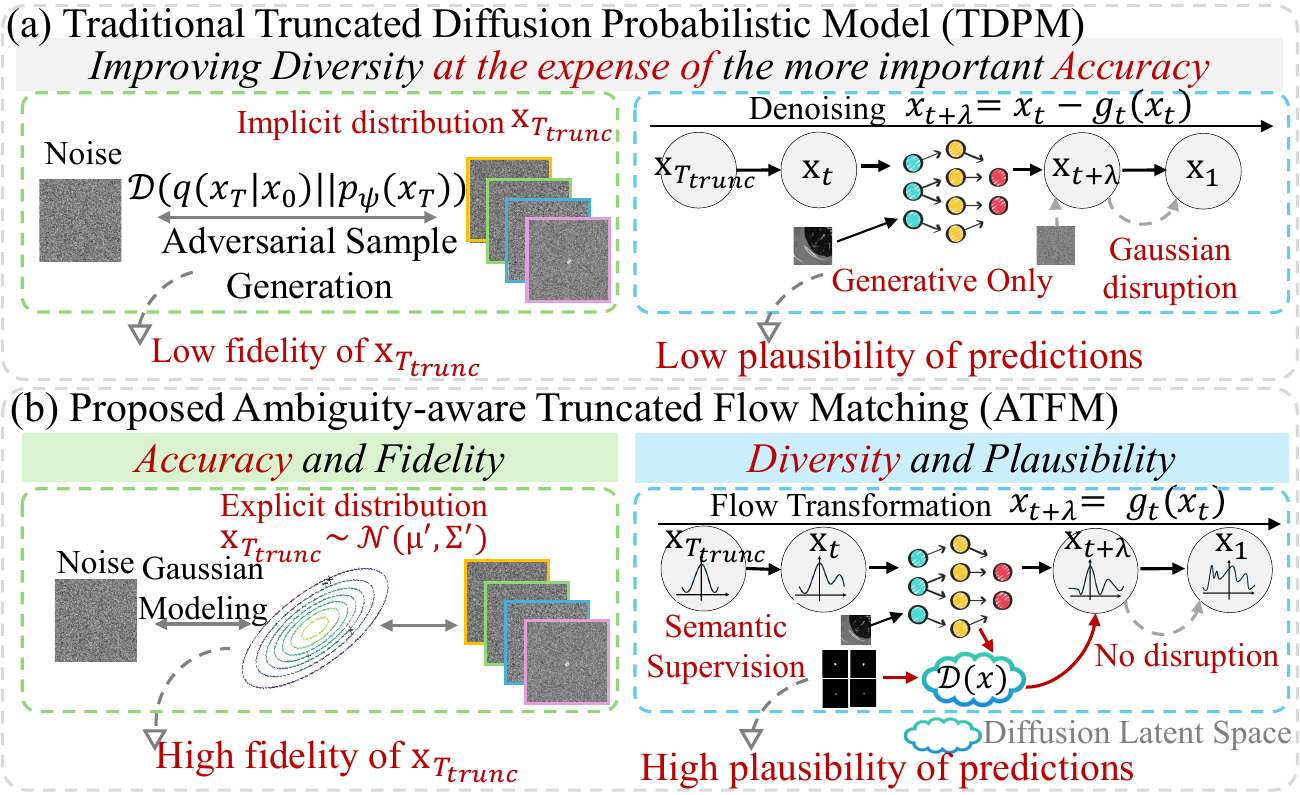} 	
	\caption{(a) Traditional TDPMs face the challenges of low fidelity and plausibility of predictions by improving diversity at the expense of accuracy. (b) The proposed ATFM enhances fidelity and plausibility of predictions by assigning distinct inference goals into two stages.} 
	\label{fig:intro2} 
\end{figure}

Truncated Diffusion Probabilistic Models (TDPMs) \cite{29} have shown great potential in simultaneously improving accuracy and diversity thanks to the inference paradigm shift. TDPMs are proposed to inference within fewer steps by leveraging an auxiliary network to model the distribution at a predefined truncation point, shifting the original inference for acceleration. TDPMs are widely used in multiple areas such as high-resolution MRSI generating \cite{30}, autonomous driving \cite{31}, remote sensing \cite{32}, demonstrating superiority in both effectiveness and efficiency of inference.

However, as Fig. \ref{fig:intro2}(a), directly applying TDPMs to achieve synergistic optimization of prediction accuracy and diversity in AMIS remains challenging due to: 
\textbf{(1) The uniform inference objective across all stages in conventional TDPMs inherently constrains their capacity for simultaneous accuracy and diversity improvement. } Vanilla TDPMs adopt a two-stage inference process with one objective mainly for acceleration. However, the deterministic and ambiguous components remain entangled throughout the inference process, making it difficult to achieve a simultaneous enhancement of both accuracy and diversity.
\textbf{(2) Sub-optimal approximation of the underlying distribution at the truncation point fundamentally compromises prediction fidelity. }  Distribution at $T_{\text{trunc}}$ is estimated by drawing samples from adversarial networks rather than explicitly modeling in vanilla TDPMs, leading to degraded fidelity caused by inconsistent predictions and omission of low-frequency modes (clinically plausible yet rare).
\textbf{(3) The absence of semantic guidance following truncation in conventional TDPMs adversely affects the plausibility of generated predictions.} Vanilla TDPMs rely on a vanilla diffusion process guided solely by generation quality after $T_{\text{trunc}}$. In AMIS tasks, this final stage lacks explicit semantic constraints for segmentation, which, although enhancing diversity, significantly compromises the more important accuracy and plausibility.

In this work, we propose the Ambiguity-aware Truncated Flow Matching (ATFM) to achieve a synergistic enhancement of accuracy and diversity of predictions in AMIS tasks, supported by three designs (as Fig. \ref{fig:intro2}(b)): 
(1) We propose Data-Hierarchical Inference, an innovative AMIS-specific inference paradigm redefinition inspired by TDPMs, where stochasticity during diffusion is marginalized for an effective disentanglement of accuracy and diversity. Specifically, ATFM performs truncated steps at the data distribution level to prioritize accuracy with diversity marginalized, whereas the final diffusion stage operates at the data sample level to enhance diversity without sacrificing accuracy.
(2) We propose Gaussian Truncation Representation (GTR), which explicitly models the Gaussian latent distribution at truncation point to enhance prediction fidelity. While traditional TDPMs approximate the implicit distribution via adversarial sampling, GTR encodes image-level semantic features to logit distribution, thereby directly modeling the reliable distribution at the truncation point. This design preserves low-frequency modes and improves consistency, leading to predictions with higher fidelity in AMIS tasks.
(3) We propose Segmentation Flow Matching (SFM), which introduces semantic-aware flow transformation to increase diversity and enhance plausibility simultaneously. While traditional TDPMs focus solely on generative quality after truncated steps, the proposed SFM leverages Flow Matching (FM) to overcome Gaussian limitations that disrupt fine-grained predictions and incorporates explicit semantic consistency modeling to ensure the plausibility of segmentation predictions in AMIS tasks.

In summary, our main contributions are as follows:
\begin{figure*} 	
	\centering 	 	
	\includegraphics[width = 1\textwidth]{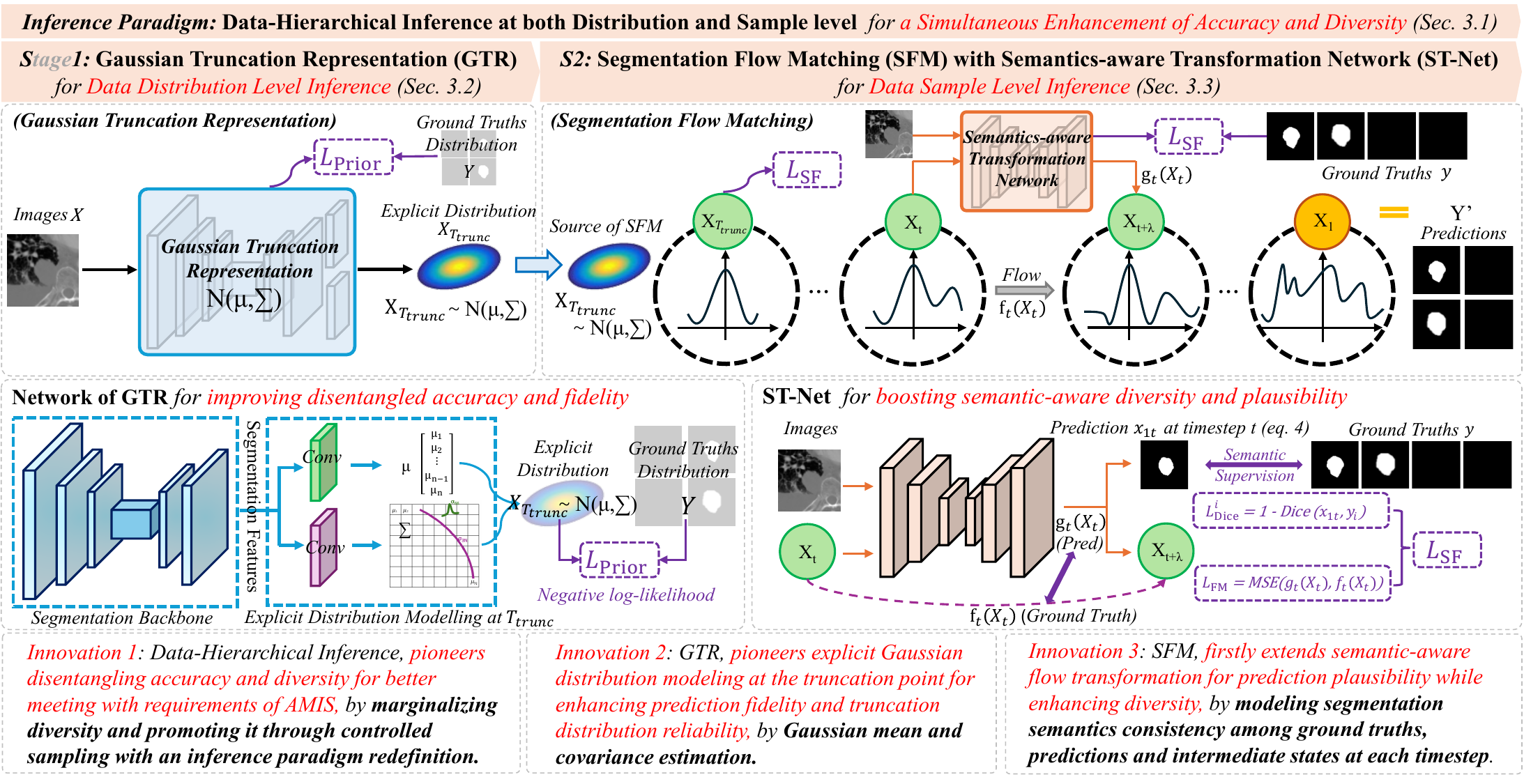} 	
	\caption{The proposed ATFM addresses the challenge of a synergistic optimization by boosting accuracy at distribution level and diversity at sample level within Data-Hierarchical Inference (Sec. 3.1) while enhancing fidelity and plausibility with GTR (Sec. 3.2) and SFM (Sec. 3.3), respectively.} 
	\label{fig:main} 
\end{figure*}
\begin{itemize}
	\item We propose ATFM with Data-Hierarchical Inference redefining a more suitable inference paradigm for AMIS for the first time, where Data-Hierarchical Inference effectively decouples accuracy and diversity by marginalizing the stochasticity during diffusion process, thereby enabling a simultaneous improvement of both.
	\item The proposed GTR in ATFM pioneers explicitly modeling a Gaussian distribution at truncation point, which effectively preserves low-frequency modes and ensures sample consistency, thereby significantly improving prediction fidelity to ground truth distribution and reliability of the distribution at truncation point.
	\item The proposed SFM in ATFM pioneers semantic-aware flow transformation by modeling semantic consistency at each timestep, thereby enhancing plausibility while emphasizing sample-wise diversity, and is built upon the Flow Matching (FM) process that inherently avoids disturbance from Gaussian constraints.
	\item A comprehensive evaluation on the LIDC and ISIC3 subset datasets demonstrates that proposed ATFM significantly improves the SOTA methods and simultaneously achieves a more efficient inference. 
\end{itemize}

\section{Related Work}

\subsection{Ambiguous Medical Image Segmentation}

Existing AMIS approaches fall into four main paradigms: model ensemble, multi-head frameworks, conditional variational autoencoder (cVAE)-based models, and diffusion-based models. All face a fundamental trade-off between prediction accuracy and sample diversity \cite{6}.

Model ensemble \cite{10,12} and multi-head models \cite{8,15} generate multiple predictions via diverse architectures or output heads but do not change the original inference process. Consequently, prediction quality heavily depends on model selection, limiting simultaneous improvement of accuracy and diversity.

CVAE-based \cite{7,16} and diffusion-based methods \cite{1,3} inject stochasticity to enhance diversity, yet both follow a one-stage inference paradigm that entangles accuracy and diversity optimization. This leads to inherent conflicts where gains in diversity often reduce accuracy.

To address the challenges, in the proposed ATFM, we introduce Data-Hierarchical Inference to redefine the inference paradigm for AMIS inspired by TDPMs. Specifically, a principled decoupling of the two objectives is achieved across different data hierarchies,  where accuracy is enhanced at distributional level and diversity is promoted at sample level.

\subsection{Truncated Diffusion Probabilistic Models}

Truncated Diffusion Probabilistic Models (TDPMs) accelerate inference by truncating the diffusion process at \( T_{\text{trunc}} \ll T \), splitting inference into estimating the distribution at \( T_{\text{trunc}} \) and reverse diffusion thereafter. Existing TDPMs approximate the distribution at \( T_{\text{trunc}} \) via adversarial sampling or perturbations~\cite{29,30}, improving speed but without redefining the inference paradigm. Consequently, they lack explicit modeling and semantic supervision at \( T_{\text{trunc}} \), limiting performance on AMIS tasks.

The proposed ATFM addresses these gaps by introducing Data-Hierarchical Inference, which marginalizes stochasticity before truncation to disentangle and enhance both accuracy and diversity. Consequently, ATFM explicitly models a Gaussian distribution at \( T_{\text{trunc}} \) for high-fidelity sampling and applies semantic supervision after truncation to improve plausibility, making ATFM a tailored solution for AMIS.

\section{Methods}

\begin{figure}
	\centering 	 	 
	\includegraphics[width = 0.47\textwidth]{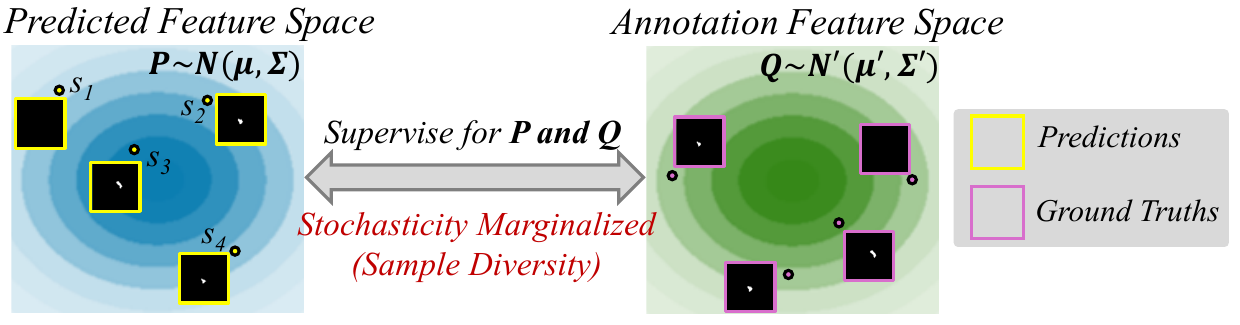} 	
	\caption{Data-Hierarchical Inference disentangles accuracy and diversity by marginalizing stochasticity during diffusion with a data-distribution level supervision.} 
	\label{fig:DHI} 
\end{figure}

The proposed ATFM (Fig. \ref{fig:main}) redefines the inference paradigm for disentanglement and simultaneous enhancement of accuracy and diversity by marginalizing stochasticity during truncation, forming an AMIS-specific defining solution. (\textit{Data-Hierarchical Inference, Sec. 3.1}). Specifically, ATFM firstly improves prediction fidelity and truncation distribution reliability by explicitly modeling the Gaussian distribution at truncation point, thereby preserving low-frequency modes often missed by sampling-based approximation and ensuring consistency across predictions (\textit{GTR, Sec. 3.2}). Secondly, ATFM extends semantic-aware flow transformation by modeling semantic consistency among labels, predictions, and intermediate states during flow matching (FM), thereby enhancing prediction plausibility while promoting diversity. Additionally, Gaussian constraints are avoided for fine-grained predictions by FM (\textit{SFM, Sec. 3.3}).

\subsection{Data-Hierarchical Inference Enables Principled Disentanglement and Joint Enhancement of Accuracy and Diversity}

The proposed Data-Hierarchical Inference forms disentanglement and simultaneous enhancement of prediction accuracy and diversity by marginalizing the stochasticity during diffusion, which introduces a redefinition of AMIS-specific inference paradigm. While preserving the efficiency gains from truncation, Data-Hierarchical Inference introduces a principled separation between distribution-level and sample-level inference, dedicated to optimizing overall accuracy and prediction diversity, respectively.

Specifically, the overall inference process firstly focuses on improving accuracy by supervising an accurate explicit distribution at $T_{\text{trunc}}$ as eq. \ref{222}, and then enhances diversity by generating varied samples from this distribution after $T_{\text{trunc}}$ as eq. \ref{333}, ensuring each prediction remains both distinct and consistent with the underlying semantics.

\begin{equation}
	\label{222}
	\underbrace{\{s_1, s_2, \dots, s_n\}}_{\text{diversity marginalized}} 
	\in
	\underbrace{P \sim \mathcal{N}(\mu, \Sigma) \stackrel{\text{supervise}}{\Longleftrightarrow} Q\sim\mathcal{N'}(\mu', \Sigma')}_{\text{distribution level for accuracy}}
\end{equation}

\begin{equation}
	\label{333}
	\underbrace{\ s_i \ \xrightarrow{\text{diffusion}} \ \text{pred}_i \, \quad
		\{ \text{pred}_i \}_{i=1}^n \stackrel{\text{supervise}}{\Longleftrightarrow} \{ \text{gt}_i \}_{i=1}^n}_{\text{sample level for diversity}}
\end{equation}
where \( P \) denotes the explicit intermediate distribution estimated by our Data-Hierarchical Inference at \( T_{\text{trunc}} \), and \( Q \) represents the corresponding distribution derived from ground truths. $s_i$ denote latent samples, $\text{pred}_i$ are the corresponding predictions, and $\text{gt}_i$ represent the ground truths.

As illustrated in Fig.~\ref{fig:DHI} and eq. \ref{222}, Data-Hierarchical Inference fundamentally redefines the inference paradigm by marginalizing stochasticity during truncation to achieve a principled disentanglement of accuracy and diversity. This paradigm optimization enables an improvement in accuracy without compromising diversity. Sample-level diffusion builds upon the globally aligned explicit distribution at $T_{\text{trunc}}$, leading to a unified and robust enhancement of both prediction fidelity and diversity.

Data-Hierarchical Inference inherently addresses the core challenges of AMIS by reconciling high prediction accuracy with plausible diversity. Through principled disentanglement of accuracy and diversity, Data-Hierarchical Inference establishes a robust and efficient solution that redefines the inference paradigm and application of TDPMs in AMIS.

\begin{figure*} [t]	
	\centering 	 	 
	\includegraphics[width = 1\textwidth]{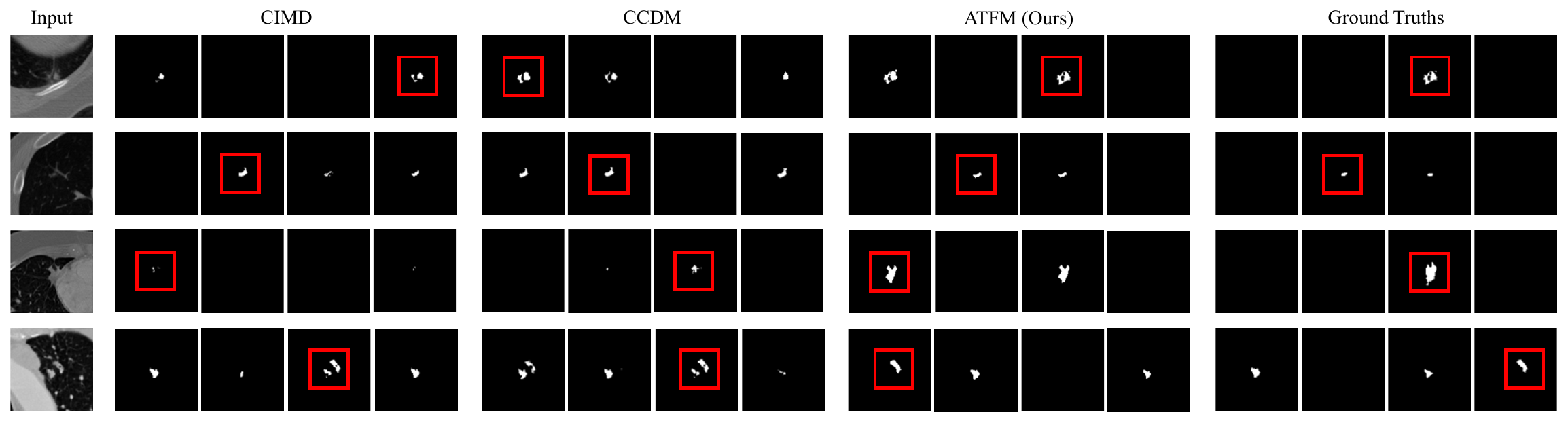} 	
	\caption{Comparative qualitative results on LIDC dataset among ground truths, two advanced methods and the proposed ATFM demonstrate both better alignment with ground truths and higher per-sample accuracy.} 
	\label{fig:visualization_comp1} 
\end{figure*}
\begin{table*}[t]      	
	\centering      		
	\begin{tabular}{l|ccc|c|cc}          			
		\toprule          			
		Methods & $\text{GED}_{16} \downarrow$ & $\text{GED}_{32} \downarrow$ & $\text{GED}_{100} \downarrow$ &  $\text{HM-IoU}_{32} \uparrow$ & $\text{MDM}_{32}$ $\uparrow$ \\
		
		\midrule          			
		Prob. Unet\cite{15} & $0.310_{\pm0.010}$ & $0.303_{\pm0.010}$ & $0.252_{\pm0.004}$ & $0.548_{\pm 0.000}$ & $0.681_{\pm0.020}$  \\

		PHiSeg\cite{7} & $0.262_{\pm0.000}$ & $0.247_{\pm0.000}$ & $0.224_{\pm0.004}$ & $0.595_{\pm0.000}$ & $0.704_{\pm0.020}$  \\  
		
		SSN\cite{10} & $0.259_{\pm0.000}$ & $0.243_{\pm0.010}$ & $0.225_{\pm0.002}$ & $0.550_{\pm0.010}$ & - \\ 
		
		MoSE\cite{20} & $0.218_{\pm0.003}$ & $0.195_{\pm0.002}$ & $0.189_{\pm0.002}$ & $0.624_{\pm0.004}$ & $0.767_{\pm0.004}$ \\
		
		P$^{\text{2}}$SAM\cite{25} & $0.208_{\pm0.000}$ & $0.206_{\pm0.000}$ & - & $0.627_{\pm0.000}$ & $0.939_{\pm0.000}$ \\
		
		CIMD\cite{1} & - & - & $0.321_{\pm0.000}$ & $0.592_{\pm0.002}$ & $0.915_{\pm0.004}$\\  
		
		AB\cite{19} & $0.213_{\pm0.001}$ & $0.196_{\pm0.002}$ & $0.193_{\pm0.002}$ & $0.619_{\pm0.001}$ & $0.792_{\pm0.002}$ \\ 
		
		CCDM\cite{3} & $0.212_{\pm0.001}$ & $0.194_{\pm0.001}$ & $0.183_{\pm0.002}$ & $0.631_{\pm0.002}$ & $0.790_{\pm0.003}$ \\
		
		ATFM(Ours) & $\mathbf{0.206_{\pm0.002}}$ & $\mathbf{0.188_{\pm0.001}}$ & $\mathbf{0.162_{\pm0.002}}$ & $\mathbf{0.667_{\pm0.002}}$ & $\mathbf{0.948_{\pm0.001}}$ \\  
		\bottomrule
		
	\end{tabular}

	\caption{Quantitative results on LIDC dataset show the superior performance of ATFM. Bold represents the best per column. Arrows indicate the increasing performance of the metrics.} 
	
	\label{tab:1}
\end{table*}

\noindent\textbf{Summarized Advantage}: Data-Hierarchical Inference, pioneers disentangling accuracy and diversity for better meeting with requirements of AMIS, by marginalizing diversity and promoting it through controlled sampling within an inference paradigm redefinition in two consecutive stages.

\subsection{Gaussian Truncation Representation Improves Fidelity via Explicit Gaussian Modeling} 
The proposed GTR models the explicit distribution at \( T_{\text{trunc}} \) for prediction fidelity by parameterizing it as a Gaussian distribution. This explicit modeling, supervised to ensure the overall accuracy and serving as the truncation step, enhances fidelity of predictions and reliability of distribution at \( T_{\text{trunc}} \) which preserves low-frequency modes and improves sample consistency compared to sampling-based approximations.

\begin{theorem}
	\label{T1}
	The marginal distribution of the latent variable at any diffusion timestep $\tau$ can be parameterized as \begin{equation}
		\mathcal{N}(\mu, \Sigma), \quad \text{with } \Sigma = D D^\top + L.
	\end{equation}
\end{theorem}

\begin{theorem}
	\label{T2}
	For any Gaussian distribution $\mathcal{N}(\mu_0, \Sigma_0)$, there exists a specific timestep $\tau^*$ at which the diffusion process produces an identical distribution.
\end{theorem}

\textit{The proof of Theorems \ref{T1} and \ref{T2} is provided in the appendix of extended version.}

According to Theorems \ref{T1} and \ref{T2}, together with the controllability of diffusion trajectories between adjacent  timesteps \cite{37}, it can be concluded that arbitrary Gaussian distributions are admissible as distribution within the diffusion framework. Hence, the Gaussian distribution on the logit map modeled as eq. \ref{3} following the formulation in Theorem \ref{T1} is selected as the truncation distribution, as it most closely approximates the predictions and enables supervision to achieve optimal accuracy and reliability. 
\begin{equation} \label{3}
	Z = f_{\theta}(X), 
	\mu = g_{\phi}(Z), 
	\Sigma = h_{\psi}(Z), 
	X_{T_{\text{trunc}}} \sim \mathcal{N}(\mu, \Sigma)
\end{equation}
where \( f_{\theta} \), \( g_{\phi} \), and \( h_{\psi} \) denote the segmentation backbone and separate convolutional layers for estimating the mean and covariance, respectively, and $X_{T_{\text{trunc}}}$ denotes the Gaussian distribution at the truncation point. 

The $L_{\text{Prior}}$ of GTR explicitly supervises accuracy between the truncation distribution and ground truths, defined as:

\begin{align} \label{loss_p} L_{\text{Prior}} &= -\log \int p(Y|X_{T_{\text{trunc}}}) p(X_{T_{\text{trunc}}}|X) dX_{T_{\text{trunc}}} \nonumber \\
	&\approx\frac{1}{M} \sum_{i=1}^{M} -\log p(Y|X_{T_{\text{trunc}}}^i) \end{align} 
where $X_{\text{trunc}}^i$ is the $i^{th}$ sample from $X_{T_{\text{trunc}}}$, $Y$ is the ground truth, and $M$ is the number of Monte Carlo samples. Minimizing the negative log-likelihood in $L_{\text{Prior}}$ optimizes the explicit Gaussian $X_0$ for accuracy and fidelity. The network is then frozen for subsequent inference.

\noindent\textbf{Summarized Advantage}: GTR, pioneers explicit Gaussian distribution modeling at the truncation point for enhancing prediction fidelity and truncation distribution reliability via mean and covariance parameterization and estimation.

\subsection{Segmentation Flow Matching Enhances Plausibility via Semantic Consistency Modeling} 
The proposed SFM extends semantic-aware flow transformation for plausibility by modeling semantic consistency among labels, predictions, and intermediate states at each timestep after \( T_{\text{trunc}} \) during FM training. It incorporates a Semantic-aware Transformation Network (ST-Net) at each timestep to ensure that flow transformation proceeds under semantic constraints. SFM aligns well with the ambiguity-resolving requirements of AMIS tasks by enhancing plausibility while promoting diversity. Moreover, by employing Flow Matching instead of DDPM, SFM avoids the Gaussian constraints that introduces disturbances in fine-grained predictions. 

Algorithm 1 is the summarized training procedure of SFM.

\begin{figure*} [t]	
	\centering 	 	 
	\includegraphics[width = 1\textwidth]{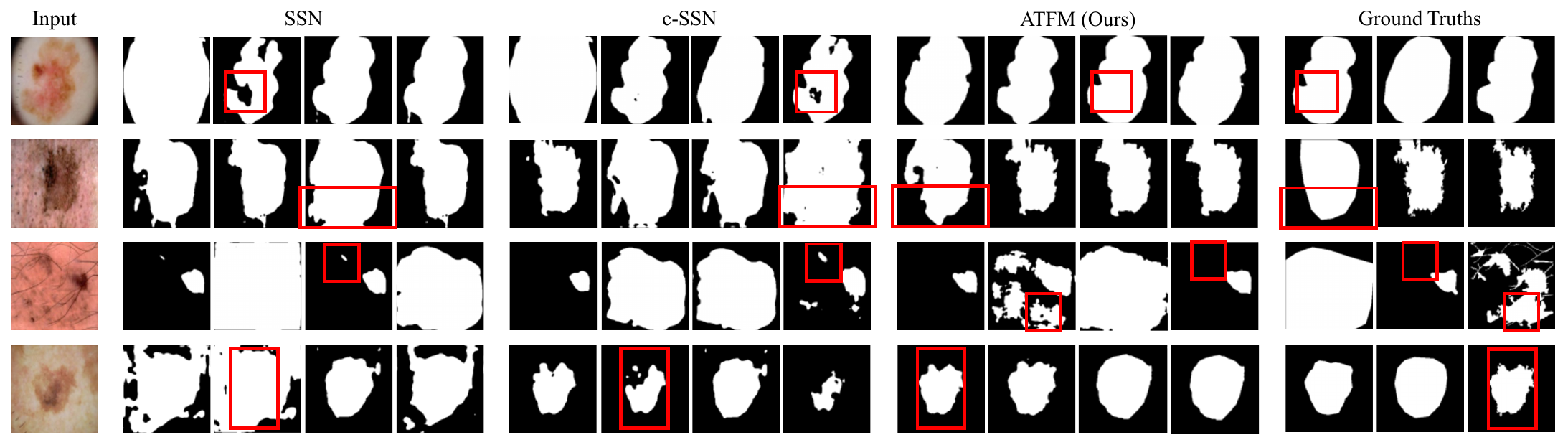} 	
	\caption{Comparative qualitative results on ISIC3 subset dataset among ground truths, two advanced methods and the proposed ATFM demonstrate both better alignment with ground truths and higher per-sample accuracy.} 
	\label{fig:visualization_comp2} 
\end{figure*}
\begin{table*}[t]
	\centering
	
	\begin{tabular}{l|ccc|c|c}          			
		\toprule          			
		Methods & $\text{GED}_{16} \downarrow$ & $\text{GED}_{32} \downarrow$ & $\text{GED}_{100} \downarrow$ &  $\text{HM-IoU}_{32} \uparrow$ & $\text{MDM}_{32}$ $\uparrow$ \\
		\midrule      		
		Prob. Unet\cite{15}  & $0.202_{\pm0.003}$ & $0.187_{\pm0.003}$ &  $0.171_{\pm0.002}$ & $0.697_{\pm0.005}$ &  $0.927_{\pm0.003}$\\ 
		SSN\cite{10} & $0.197_{\pm0.007}$ & $0.181_{\pm0.004}$  &  $0.167_{\pm0.002}$ & $0.700_{\pm0.004}$ & $0.939_{\pm0.001}$\\ 			
		c-Prob. Unet\cite{17} & $0.208_{\pm0.004}$ & $0.202_{\pm0.005}$  & $0.179_{\pm0.002}$ &  $0.719_{\pm0.004}$ & $0.925_{\pm0.004}$\\ 			
		c-SSN\cite{17} & $0.200_{\pm0.007}$ & $0.195_{\pm0.007}$  & $0.177_{\pm0.001}$  & $0.725_{\pm0.004}$ & $0.931_{\pm0.003}$\\ 			
		ATFM(Ours)  & $\mathbf{0.183_{\pm0.001}}$ & $\mathbf{0.152 _{\pm0.002}}$ & $\mathbf{0.147_{\pm0.003}}$ & $\mathbf{0.732_{\pm0.003}}$ & $\mathbf{0.942_{\pm0.002}}$\\
		\bottomrule
	\end{tabular} 
	\caption{Quantitative results on ISIC3 subset dataset show the superiority of ATFM. Bold represents the best per column. Arrows indicate the increasing performance of the metrics.}
	
	\label{tab:2}
\end{table*}


\begin{algorithm}
	\caption{Training Procedure for proposed SFM}
	\label{SFM}
	\begin{algorithmic}[1]
		\REQUIRE Source distribution \( X_{T_{\text{trunc}}} \), Target distribution $X_1$, Output of ST-net $g_{\theta}(X_t)$ at timestep t, ground truths ${y_i}$
		\REPEAT
		\STATE $\label{44} X_t = t \times X_1 + (\mathbf{1}-t) \times  X_{T_{\text{trunc}}}$ 
		\STATE $\label{55} x_{1t} = x_t + g_{\theta}(X_t) \times (\mathbf{1}-t)$ (\textit{calculation of prediction corresponding to timestep t})
		\STATE $\label{66} L_{\text{Dice}}^i = 1 - Dice(x_{1t}, y_i) , i = 1, 2, \dots, N$ 
		\STATE $\label{77} L_{\text{SF}} = L_{\text{FM}} + \frac{1}{N} \sum_{i=1}^{N} \alpha \times L_{\text{Dice}}^i $ (\textit{semantic consistency modeling})
		\STATE $\label{88} \theta \leftarrow \theta - \eta\nabla_{\theta} L_{\text{SF}} $ (\textit{gradient update})
		\UNTIL{$||\nabla_{\theta} L_{\text{SF}}|| < \delta$ (\textit{convergence})}
	\end{algorithmic}
\end{algorithm}

\textbf{Computing the intermediate prediction corresponding to timestep t (line \ref{44} and \ref{55} in Algorithm 1):} The flow transformation follows an Optimal Transformation (OT) schedule \cite{14}, representing the shortest path between source and target distributions. Under the OT framework, the diffusion trajectory in the latent space forms a line segment.
Therefore, we perform analytic geometry in the latent space: the intermediate state \( X_t \) at timestep \( t \) is computed by linear interpolation between the source endpoint \( X_{T_{\text{trunc}}} \) and the target endpoint \( X_1 \). Then, using the direction vector of the segment $g_t(X_t)$ and the position of \( X_t \), a predicted result \( x_{1t} \) is derived by projecting along the diffusion trajectory starting from \( x_t \).

\textbf{Semantic consistency modeling (line \ref{66} to \ref{88} in Algorithm 1):} By computing the Dice loss between the predicted result \( x_{1t} \) and all ground truth annotations, semantic consistency at timestep \( t \) can be explicitly modeled. This supervision acts as an auxiliary constraint to the Flow Matching loss, encouraging the transformation to preserve plausibility and consistency throughout the diffusion process for diversity.

The aforementioned SFM training process not only ensures accurate flow transformation, but also explicitly models the semantic consistency among the state, predicted result and ground truths at each timestep \( t \). This dual-objective optimization enhances the semantic plausibility of predictions, and simultaneously capturing diverse sample-level variations via flow matching, positioning SFM as an indispensable module of the proposed ATFM framework for AMIS.

\noindent\textbf{Summarized Advantage}: SFM, firstly extends semantic-aware flow transformation for prediction plausibility while enhancing diversity, by modeling segmentation semantic consistency among ground truths, predictions and intermediate states at each timestep.

\section{Experiments}
\subsection{Experimental Setup}
\subsubsection{Datasets.} In our experiments, we applied two public datasets for ambiguous medical image segmentation: LIDC-IDRI \cite{23} and ISIC3 subset \cite{24,17}. The LIDC-IDRI dataset consists of lung CT scans with multiple expert-annotated lesion segmentations, highlighting diagnostic ambiguities. Following preprocessing as described in \cite{15,16}, the dataset includes 15,096 slices, each with four corresponding segmentation labels. The ISIC3 dataset provides dermoscopic images of skin lesions with annotations for lesion boundaries. Using the preprocessed ISIC3 subset from \cite{17}, we work with 300 images, each featuring exactly three annotations.

\begin{table}
	
	\centering
	\begin{tabular}{l|c}          			
		\toprule          			
		Models & $\text{Inference Steps and Time}_{100}$ \\
		\midrule      		
		CIMD & $T = 100$ steps $\approx$ 420s \\ 
		AB & $ T = 250$ steps $\approx$ 1050s\\ 			
		CCDM & $ T = 250$ steps $\approx$ 1100s\\ 
		ATFM & GTR + ( $ T_{\text{Trunc}} $ = 25 steps) $\approx$ \textbf{113s}\\
		\bottomrule
	\end{tabular}
	\caption{Time comparison for generating 100 samples on the LIDC dataset for diffusion-based methods demonstrates the superior time efficiency of proposed ATFM.}
	
	\label{tab:4}
\end{table}

\subsubsection{Implementation Details.}
All training and inference procedures are conducted on a single RTX 3090 GPU with 24GB memory. SFM in ATFM is trained for 200 epochs on LIDC with GTR pretrained for 1000 epochs, and for 120 epochs on ISIC3 with a 400-epoch GTR. We set $\lambda = 10^{-3}$ (i.e. $T=1000$) with a linear schedule for all experiments. Both GTR and ST-Net of the SFM are optimized using an Adam optimizer \cite{28} with a learning rate of $10^{-4}$. Hyper-parameter $M$ is set to $20$ following \cite{17} and $\alpha$ is set to $10^{-3}$ and $10 ^ {-4}$ for LIDC and ISIC3 respectively according to hyper-parameter studies.  

\begin{figure*} [t]
	\centering 	 	
	\includegraphics[width = 1\textwidth]{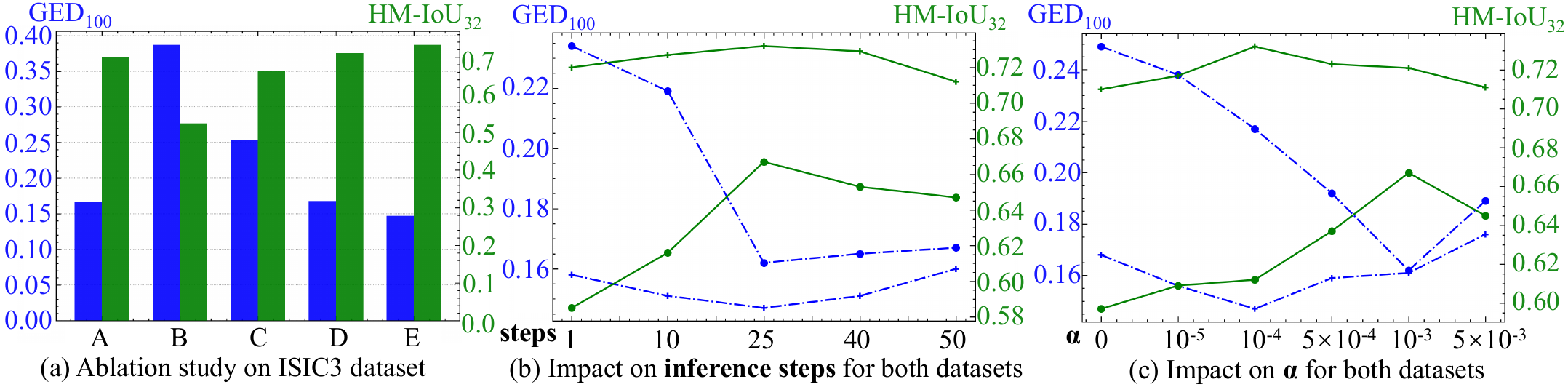} 	
	\caption{Comprehensive analysis through ablation and hyper-parameter studies demonstrates the effectiveness of innovations and training configurations in proposed ATFM.} 
	\label{fig:ablation} 
\end{figure*}

\begin{table}
	
	\centering
	{\rmfamily  
		\begin{tabular}{l|c|c}          			
			\toprule          			
			Models & $\text{GED}_{100} \downarrow$ &  $\text{HM-IoU}_{32}\uparrow$ \\
			\midrule      		
			Act. GTR & $0.230_{\pm0.001}$ & $0.550_{\pm0.010}$ \\ 
			SFM w/o $L_{SF}$  & $0.185_{\pm0.002}$ & $0.624_{\pm0.002}$ \\ 			
			SFM   & $0.176_{\pm0.002}$ & $0.631_{\pm0.002}$ \\ 
			ATFM w/o $L_{SF}$& $0.249_{\pm0.001}$  & $0.597_{\pm0.003}$ \\ 			
			ATFM & $\mathbf{0.162_{\pm0.002}}$ &  $\mathbf{0.667_{\pm0.002}}$ \\
			\bottomrule
		\end{tabular}
	}
	\caption{Ablation Study on LIDC dataset shows the validity of all components in proposed ATFM.}
	
	\label{tab:3}
\end{table}
\subsubsection{Evaluation Metrics.}
Three metrics are utilized for comprehensive evaluation from three aspects: For segmentation distribution, we utilize the Generalised Energy Distance (GED) \cite{18} to evaluate the alignment among distribution of predictions and ground-truths. For sample fidelity, we utilize the Hungarian-Matching Intersection-over-Union (HM-IoU) \cite{20} to provide an accurate representation of the performance on segmentation across all predictions. For individual segmentation accuracy, we utilize the Maximum Dice Matching (MDM) \cite{1} to evaluate the best Dice scores between each prediction result and each ground truth. We denote the metrics with subscript n to represent the metrics calculated using n samples. Results are reported as mean ± standard deviation over five independent runs.

\subsection{Experimental Results}

\subsubsection{Comparison with SOTAs for Performance Superiority.} 

\paragraph{Quantitative Evaluation.} Table \ref{tab:1} and Table \ref{tab:2} report the quantitative results on the LIDC and ISIC3 datasets, respectively. Across all key metrics—GED (for diversity), HM-IoU (for sample fidelity), and MDM (for individual accuracy)—ATFM consistently outperforms state-of-the-art methods. Notably, ATFM achieves an 11.5\% improvement on GED$_{100}$ in LIDC, along with consistent gains in GED$_{16}$ and GED$_{32}$ compared with the runner-up method, highlighting its ability to better capture the underlying segmentation label distribution. A minimum of 7.3\% improvement in HM-IoU$_{32}$ further demonstrates the superior fidelity of the generated samples. ATFM also leads in the MDM$_{32}$ metric, validating its accuracy at the individual prediction level. Similar trends are observed on ISIC3, where ATFM outperforms the runner-up method by 12\% in GED while also achieving top-tier results in HM-IoU and MDM, demonstrating comprehensive improvements in diversity, fidelity, and accuracy. These results showcase the effectiveness of the Data-Hierarchical Inference framework of ATFM, which explicitly models intermediate distributions and enforces semantic-aware flow transformations to enhance accuracy and diversity in AMIS.

\paragraph{Qualitative Results.} Fig. \ref{fig:visualization_comp1} and Fig. \ref{fig:visualization_comp2} further compare visual results among advanced methods and proposed ATFM for LIDC dataset and ISIC3 dataset, respectively. Predictions from ATFM more faithfully reflect the range of plausible annotations and better preserve fine-grained structures, indicating its superior alignment with ambiguous and detailed ground truths, which is an essential and necessary requirement in AMIS tasks.

\paragraph{Inference Efficiency.} Table \ref{tab:4} reports the total inference time for generating 100 plausible predictions. ATFM retains the inference efficiency advantage of truncated diffusion models, requiring only 25 diffusion steps and an estimation of truncation point (approx. 113s). Compared to other diffusion-based methods, it achieves both superior segmentation performance and significantly faster inference, highlighting its practicality and scalability in AMIS applications.

\subsubsection{Ablation Studies Demonstrate Effectiveness of Innovations.} Table \ref{tab:3} and Fig. \ref{fig:ablation}(a) show the conducted ablation study on both datasets evaluating five model variants: GTR with activation layers (A), SFM with and without \(L_{\text{SF}}\) (B and C), and ATFM with and without \(L_{\text{SF}}\) (D and E). ATFM outperformed both Act. GTR and SFM by a minimum of $10\%$ and $6\%$ on both metrics, highlighting the benefit of Data-Hierarchical Inference and the high effectiveness and fidelity of proposed ATFM provided by GTR. The performance gap of an average of $11\%$ between models with and without \(L_{\text{SF}}\) emphasizes the importance of semantics consistency modeling, underscoring the role of \(L_{\text{FM}}\) and SFM in preserving plausibility when enhancing diversity.

\subsubsection{Hyper-parameter Studies Prove Effectiveness of Training Configurations.} Fig. \ref{fig:ablation}(b) and \ref{fig:ablation}(c) illustrate the effects of inference step count in SFM and \( \alpha \) in \( L_{\text{SF}} \) in SFM, respectively. Five values between 1 and 50 are set for inference steps based on the property of Euler Sampler \cite{21,22}. Optimal performance was achieved within 25 steps, offering a good balance between performance and efficiency. For \( \alpha \), values between 0 and \( 5 \times 10^{-3} \) were tested. Small $\alpha$ values limits \( L_{\text{SF}} \)'s impact, while $\alpha$ with too large values diminishes the effect of \( L_{\text{FM}} \). Values of $\alpha$ striking a balance were chosen as the final setting.

\section{Conclusion}

In this work, we proposed Ambiguity-aware Truncated Flow Matching (ATFM), addressing the challenge of jointly improving prediction accuracy and diversity via a novel inference paradigm and dedicated model components, tailored to the demands of AMIS tasks. Specifically, we firstly proposed Data-Hierarchical Inference, redefining a novel inference paradigm that disentangles prediction accuracy and diversity, which supervises a distribution for accuracy at $T_{\text{trunc}}$ by marginalizing stochasticity and promoting diversity through controlled sampling in the following timesteps. On this foundation, we designed two key modules for ATFM: GTR, which explicitly models the Gaussian distribution at truncation point to ensure prediction fidelity and truncation distribution reliability for overall accuracy; and SFM, which extends semantic-aware flow transformation to model semantic consistency across predictions, annotations and intermediate states for enhancing prediction plausibility while promoting diversity. Experimental results on two public datasets showed that the proposed ATFM outperforms SOTA methods across all metrics and offers a more efficient inference process simultaneously. ATFM offers a versatile and reliable solution for AMIS across a broader range of scenarios through multifaceted analysis and outstanding performance.

\section{Acknowledgments}
This work was supported by the National Natural Science Foundation of China under Grants 62501195, 62272135, 62372135 and 82527807, and the Key Research \& Development Program of Heilongjiang Province under Grants 2023ZX01A08 and 2024ZX12C23, and the Natural Science Foundation of Heilongjiang Province under Grant LH2024F019.

\section{Appendix 1: Discussion of the Underlying Mechanism}

\begin{figure} [H]
	\centering 	 
	\includegraphics[width = 0.48\textwidth]{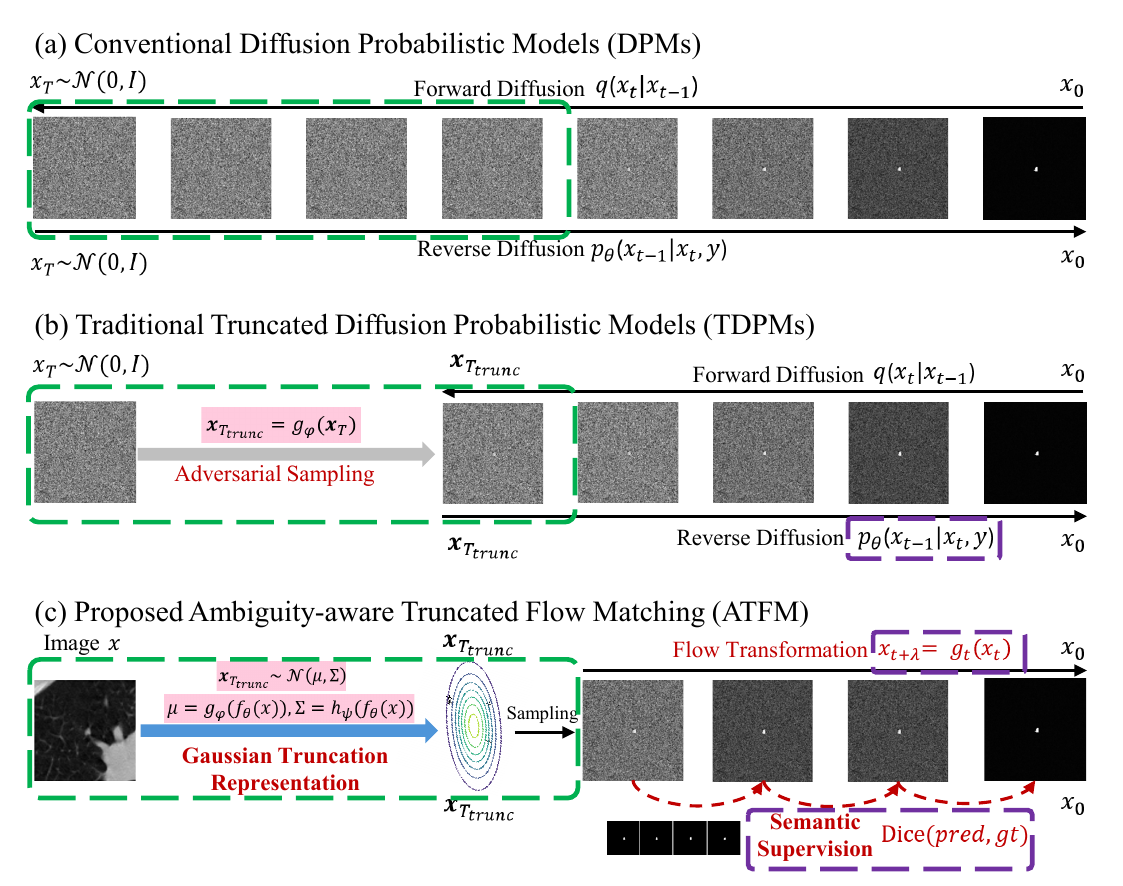} 	
	\caption{The comparison of inference paradigms among DPMs, TDPMs, and ATFM (1) reveals that the inference paradigm shift is the underlying mechanism which makes TDPMs potential for ambiguous medical image segmentation (AMIS), and (2) highlights the task-specific improvements introduced by ATFM for the AMIS scenario.} 
	\label{xxx}	
\end{figure}

\begin{figure*}
	\centering 	 	
	\includegraphics[width = 1\textwidth]{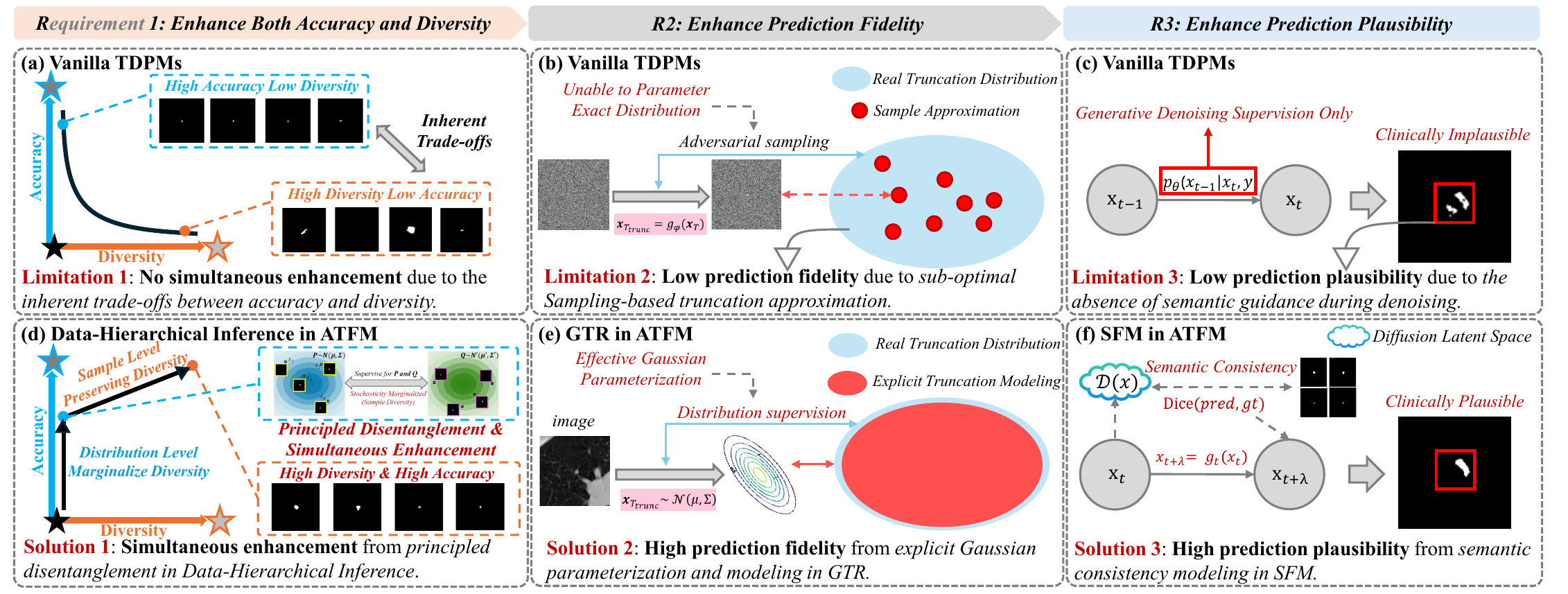} 	
	\caption{Limitations of vanilla TDPMs and the solutions offered by the proposed ATFM. From R1 to R3, they are the necessary requirements of performing reliable AMIS. From (a) to (c), they
		are the limitations of vanilla TDPMs, and from (d) to (f), they are the solutions and advantages of ATFM, respectively. (a) Vanilla TDPMs can't achieve simultaneous enhancement of accuracy and diversity due to the inherent trade-off. (b) Vanilla TDPMs lack prediction fidelity due to sub-optimal sampling-based truncation distribution approximation. (c) Vanilla TDPMs lack prediction plausibility due to the absence of semantic guidance. (d) Data-Hierarchical Inference in proposed ATFM introduces simultaneous enhancement of accuracy and diversity by a principled disentanglement. (e) GTR in proposed ATFM achieves high prediction fidelity via explicit Gaussian parameterization and modeling. (f) SFM in proposed ATFM achieves high prediction plausibility from semantic modeling in SFM.} 
	\label{CS} 
\end{figure*}

\section{Appendix 2: Potential Underlying Mechanism of Truncated Diffusion Probabilistic Models}
Truncated Diffusion Probabilistic Models (TDPMs) demonstrate their potential through the introduction of an inference paradigm shift—a concept that optimizes the inference trajectory between start point and a predefined truncation timestep $T_{\text{trunc}}$ of diffusion models. Specifically, TDPMs retain the original starting point (random noise) and truncation point ($X_{T_{\text{trunc}}}$) of the diffusion process, while strategically replacing the intermediate inference steps between $t=0$ and $t=T_{\text{trunc}}$ with a learned truncation module (usually adversarial sampling). This substitution constitutes a paradigm shift, proving that the intermediate reasoning path of diffusion models is not rigid but can be restructured. (\textit{As the comparison illustrated in green boxes in Fig. \ref{xxx}(a) and Fig. \ref{xxx}(b)})

Although the primary goal of TDPMs is to accelerate inference, their architecture implicitly reveals a structural flexibility in diffusion-based generative models. By truncating and replacing part of the denoising trajectory, TDPMs validate the feasibility of modeling this process differently. This foundational insight is particularly valuable for ambiguous medical image segmentation (AMIS), where jointly improving prediction accuracy and diversity remains a key challenge. Thus, TDPMs not only serve as a faster alternative, but also inspire a reconsideration of how inference is formulated in diffusion frameworks especially for AMIS tasks.

\section{Appendix 3: Superior Mechanism of the proposed Ambiguity-aware Truncated Flow Matching for AMIS Inspired by TDPMs}
Inspired by the inference paradigm shift introduced by TDPMs, the proposed ATFM redefines an AMIS-specific inference paradigm. ATFM further reformulates not only the overall inference paradigm, but the detailed truncation process and the remaining diffusion steps as well to address the inherent challenges of AMIS. (\textit{as the improvement illustrated in Fig. \ref{xxx}(b) and Fig. \ref{xxx}(c)})

Firstly, the overall inference paradigm is redefined through the proposed Data-Hierarchical Inference, which assigns distinct objectives to different stages. The truncation step focuses on enhancing overall accuracy at the distribution level without compromising diversity, while the subsequent diffusion stage promotes sample-level diversity based on the high-fidelity distribution obtained earlier. This overall design enables a principled disentanglement and simultaneous improvement of both accuracy and diversity. (\textit{overall comparison between Fig. \ref{xxx}(b) and Fig. \ref{xxx}(c)})

Secondly, the truncation stage is redefined via the proposed Gaussian Truncation Representation (GTR), which explicitly models the truncation point as a learnable Gaussian distribution with parameterized mean and covariance. This replaces the sampling-based approximation in TDPMs with a stable and semantically meaningful representation. As a result, it enhances prediction fidelity and provides a reliable initialization for subsequent inference. (\textit{comparison illustrated in green boxes between Fig. \ref{xxx}(b) and Fig. \ref{xxx}(c)})

Thirdly, the post-truncation diffusion stage is redefined through Segmentation Flow Matching (SFM), which supervises semantic consistency between intermediate states, predictions, and ground truths across timesteps. This transforms the diffusion process from unconstrained denoising into a semantics-aware transformation. FM further avoids disturbance from Gaussian constraints. SFM ensures prediction plausibility while capturing diverse sample-level variations in a structured manner. (\textit{comparison illustrated in purple boxes between Fig. \ref{xxx}(b) and Fig. \ref{xxx}(c)})

Fig. \ref{CS} further analyzes the three main limitations of vanilla TDPMs in the context of AMIS requirements, along with the corresponding solutions introduced by ATFM, which fundamentally explains the reason behind the differences between the two inference paradigms and highlights the advantages of ATFM. Specifically, these limitations relate to the overall inference paradigm, the truncation process, and the post-truncation diffusion stage. In response, ATFM addresses these issues through the proposed Data-Hierarchical Inference, Gaussian Truncation Representation, and Semantic Flow Matching, respectively.

\section{Appendix 4: Proof for Theorems}
\paragraph{Theorem 1:}The marginal distribution of the latent variable at any diffusion timestep $\tau$ can be parameterized as \begin{equation}
		\mathcal{N}(\mu, \Sigma), \quad \text{with } \Sigma = D D^\top + L.
	\end{equation}

\begin{proof}
	At any diffusion timestep $\tau$ with $\Sigma^*$ as the destination covariance, the latent variable of forward diffusion satisfies:
	\begin{equation}
		z_\tau \sim \mathcal{N}(\mu, \alpha(\tau)\Sigma^* + (1 - \alpha(\tau))I), \quad \alpha(\tau) = e^{-\int_0^\tau \beta(s) ds}.
	\end{equation}
	
	By setting $D D^\top = \alpha(\tau)\Sigma_0$ following Cholesky factorization \cite{36} and $L = (1 - \alpha(\tau))I$, we obtain that:
	\begin{equation}
		z_\tau \sim \mathcal{N}(\mu, D D^\top + L),
	\end{equation}
	which shows that the distribution shares the same structural form at any $\tau$, enabling universal expressivity over time.
\end{proof}

\paragraph{Theorem 2:}For any Gaussian distribution $\mathcal{N}(\mu_0, \Sigma_0)$, there exists a specific timestep $\tau^*$ at which the diffusion process produces an identical distribution.

\begin{proof}
	To prove exact matchability at some $\tau^*$, suppose the true segmentation map is $\mu_0$, and $\mu = \sqrt{\alpha(\tau)}\mu_0$. For full match in mean and covariance, we require:
	\begin{equation}
		\begin{cases}
			\sqrt{\alpha(\tau)} \mu_0 = \mu \\
			1 - \alpha(\tau) = \dfrac{\|\Sigma_0\|_F}{\|\mu_0\|_2^2}
		\end{cases}
	\end{equation}
	Letting $f(\tau) = 1 - \alpha(\tau) - \dfrac{\|\Sigma_0\|_F}{\|\mu_0\|_2^2}$, and noting $f(0) < 0$, $f(T) > 0$, Intermediate Value Theorem \cite{35} guarantees the existence of a $\tau^* \in (0, T)$ such that $f(\tau^*) = 0$.
\end{proof}

\section{Appendix 5: Detailed Derivation and Analysis of $x_{1t}$ from line 3 in Algorithm 1}

\begin{figure} [!htbp]
	\centering 	 	
	\includegraphics[width = 0.48\textwidth]{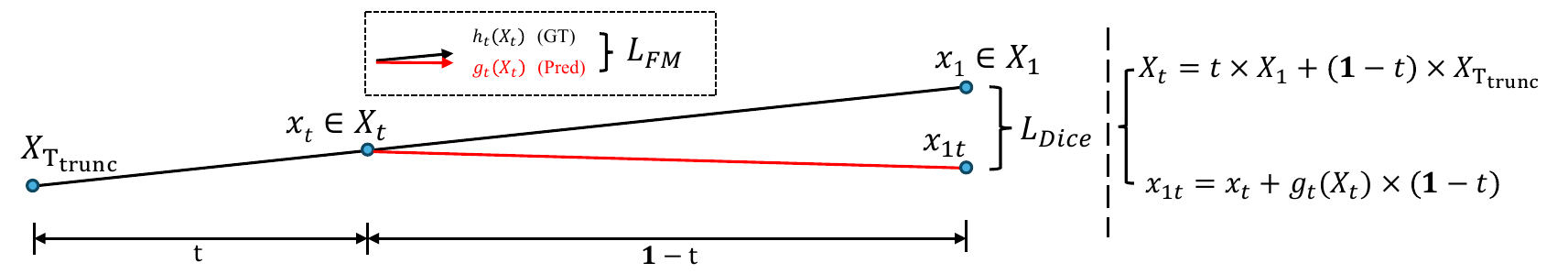} 	
	\caption{Derivation of $x_{1t}$ models the semantics consistency during flow transformation of proposed ATFM.} 
	\label{fig:X1t} 
\end{figure}

\begin{proof}
	As shown in Fig. \ref{fig:X1t}, Let \( X_{T_{\text{trunc}}} \) denote the source representation after truncation, and \( X_1 \) the final target representation in the latent space. Under the Optimal Transformation defined in the proposed ATFM, the semantic transition from \( X_{T_{\text{trunc}}} \) to \( X_1 \) forms a linear segment. For any timestep \( t \in [T_{\text{trunc}}, T] \), the latent state \( X_t \) can thus be computed as a linear interpolation:
	\[
	X_t = (1 - \alpha_t) X_{T_{\text{trunc}}} + \alpha_t X_1, \quad \text{where } \alpha_t = \frac{t - T_{\text{trunc}}}{T - T_{\text{trunc}}}.
	\]
	
	Let \( x_t \in X_t \) denote the stochastic sample at timestep \( t \). The Semantic-Transformation Network (ST-Net) is designed to predict the direction of transformation:
	\[
	g_t(X_t) \approx X_1 - X_{T_{\text{trunc}}}.
	\]
	
	With \( x_t \) and the estimated direction vector \( g_t(X_t) \), we can compute an intermediate prediction \( x_{1t} \) as:
	\[
	x_{1t} = x_t + \beta_t \cdot g_t(X_t), \quad \text{where } \beta_t = \frac{T - t}{T - T_{\text{trunc}}}.
	\]
	
	Substituting the approximation for \( g_t(X_t) \), we obtain:
	\[
	x_{1t} = x_t + \left( \frac{T - t}{T - T_{\text{trunc}}} \right) (X_1 - X_{T_{\text{trunc}}}).
	\]
	
	Following the setting of the original flow matching (FM), we set $T = 1$ and $T_{\text{trunc}} = 0$, then we can derive line 3 in Algorithm 1.
\end{proof}

As \( t \to T \), \( x_{1t} \to X_1 \), and during training, \( x_{1t} \) is supervised toward the ground truth. This modeling explicitly encodes semantic consistency during the flow transformation, ensuring that plausible and coherent predictions can still be generated even under stochastic sampling—thereby balancing diversity and accuracy within the ATFM framework.

\section{Appendix 6: Detailed Network Structures}

\subsection{Structure of Network in GTR.}

The network in GTR adopts a standard encoder-decoder architecture with 4 resolution levels. The encoder consists of 4 convolutional blocks with increasing filter sizes of 32, 64, 128, and 192. Each block applies two consecutive $3{\times}3$ convolutions followed by ReLU activations and downsampling via $2{\times}2$ max pooling. The decoder mirrors the encoder using transposed convolutions for upsampling and skip connections from corresponding encoder layers. On top of the final feature map, three separate $1{\times}1$ convolutional layers are applied to estimate the mean $\mu$, and the variance $\Sigma = DD^T + L$ for a multivariate Gaussian distribution with rank $r{=}10$, which parameterizes the explicit Gaussian distribution at $T_{\text{trunc}}$.

\subsection{Structure of ST-Net in SFM.}

ST-Net is a temporal-conditional U-Net specifically designed as the backbone of SFM in the proposed ATFM. The network adopts a four-level encoder-decoder structure with skip connections, comprising 15 residual blocks in total. Each resolution stage includes group normalization and Swish activation, followed by linear attention blocks at all levels and full self-attention at the bottleneck layer to capture both local and global dependencies.

Temporal information is injected into the network via sinusoidal time-step embeddings, which are processed through a two-layer MLP and fused into each residual block. Optional self-conditioning is supported to enhance performance. Spatial downsampling and upsampling are implemented using strided and transposed convolutions, respectively, maintaining the spatial structure of segmentation maps.

This design enables expressive and time-aware feature representation, which is essential for accurately modeling the semantic transformation process in diffusion-based ambiguous image segmentation.

\section{Appendix 7: Details of Evaluation Metrics}

\subsection{Generalised Energy Distance.} 
Generalised Energy Distance (GED) is a widely used metric for evaluating both the fidelity and diversity of segmentation predictions in ambiguous medical image segmentation tasks. It measures the discrepancy between the distributions of predicted and ground-truth segmentations. Given a set of predictions \( P = \{p_1, \dots, p_n\} \) and ground-truth annotations \( G = \{g_1, \dots, g_m\} \), the GED is computed as:
\begin{align}
	\label{ged}
	\text{GED}(P, G) = \; & 2 \, \mathbb{E}_{p \sim P,\, g \sim G} \left[d(p, g)\right] \nonumber\\
	& - \mathbb{E}_{p, p' \sim P} \left[d(p, p')\right] \nonumber \\
	& - \mathbb{E}_{g, g' \sim G} \left[d(g, g')\right]
\end{align}
where \( d(p, g) = 1 - \text{IoU}(p, g) \) denotes the dissimilarity between segmentation masks \( p \) and \( g \). A lower GED indicates that the predicted distribution more closely matches the ground-truth distribution, capturing both sample-level accuracy and overall diversity.

\subsection{Hungarian-Matching Intersection over Union.} Hungarian-Matching Intersection over Union (HM-IoU) is adopted to measure the consistency between multiple predicted segmentations and multiple ground-truth annotations. Given a set of \( C \) predictions \( P = \{p_1, \dots, p_C\} \) and ground-truth segmentations \( G = \{g_1, \dots, g_C\} \), we first compute an \( C \times C \) IoU matrix \( M^{C \times C} \) where each element is \( M_{ij} = \text{IoU}(p_i, g_j) \). The optimal one-to-one assignment \( \pi^* \) is then obtained using the Hungarian algorithm to maximize the total IoU. The HM-IoU, which offers a fair and permutation-invariant evaluation of segmentation quality across multiple predictions, is finally computed as:
\begin{equation}
	\text{HM-IoU}(P, G) = \frac{1}{C} \sum_{(i, j) \in \pi^*} \text{IoU}(p_i, g_j)
\end{equation}

For implementation, we set \( C = \mathrm{LCM}(n, m) \), where \( n \) and \( m \) represent the exact numbers of predictions and ground-truth annotations in the experiment, respectively, and \(\mathrm{LCM}\) denotes the least common multiple. This choice facilitates better alignment between the prediction and ground-truth sets.

\subsection{Maximum Dice Matching.}
Maximum Dice Matching (MDM) is used to evaluate individual segmentation accuracy for each annotation considering all predictions. Given a set of predictions \( P = \{p_1, \dots, p_n\} \) and ground-truth segmentations \( G = \{g_1, \dots, g_m\} \), MDM computes the Dice similarity coefficient between each ground truth \( g_j \) and all predictions \( p_i \). For each ground truth \( g_j \), the maximum Dice score over all predictions is selected. The overall MDM score is then obtained by averaging these maximum values across all ground truths, formally expressed as:

\begin{equation}
	\text{MDM}(P, G) = \frac{1}{m} \sum_{j=1}^m \max_{1 \leq i \leq n} \text{Dice}(p_i, g_j).
\end{equation}

This metric emphasizes the best-matching quality for each ground-truth instance, reflecting the confidence and reliability of diagnosis provided by the predictions.

\begin{figure*}
	\centering 	 	
	\includegraphics[width = 1\textwidth]{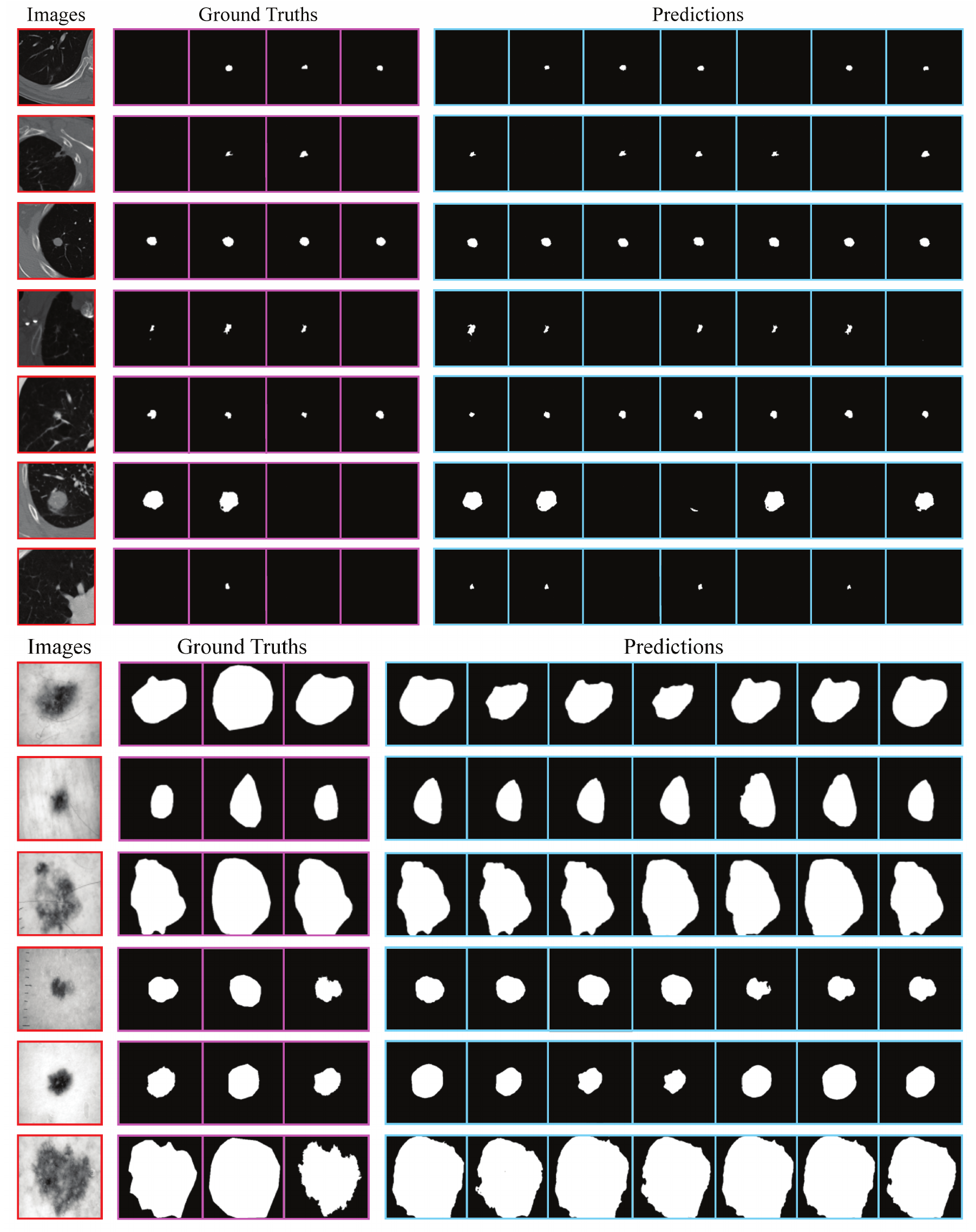} 	
	\caption{More visualization results of predictions on both datasets demonstrate effectiveness of proposed ATFM.} 
	\label{fig:Sup. LIDC} 
\end{figure*}

\section{Appendix 8: More Qualitative Results}
Fig. \ref{fig:Sup. LIDC} shows more qualitative segmentation predictions generated by the proposed ATFM and the corresponding ground-truths. The high accuracy and diversity of predictions demonstrate the superiority of the proposed ATFM.

\bibliography{mybibliography}

\end{document}